# Hierarchical Transformer Network for Utterance-level Emotion Recognition

QingBiao Li[1], ChunHua Wu[1*], KangFeng Zheng[1] and Zhe Wang[1]
[1]Beijing University of Posts and Telecommunications, Beijing, China
{liqingbiao, wuchunhua}@bupt.edu.cn, zkf_bupt@163.com, wangxiaozhe@bupt.edu.cn

## Abstract

While there have been significant advances in detecting emotions in text, in the field of utterance-level emotion recognition (ULER), there are still many problems to be solved. In this paper, we address some challenges in ULER in dialog systems. (1) The same utterance can deliver different emotions when it is in different contexts or from different speakers. (2) Long-range contextual information is hard to effectively capture. (3) Unlike the traditional text classification problem, this task is supported by a limited number of datasets, among which most contain inadequate conversations or speech. To address these problems, we propose a hierarchical transformer framework (apart from the description of other studies, the "transformer" in this paper usually refers to the encoder part of the transformer) with a lower-level transformer to model the word-level input and an upper-level transformer to capture the context of utterance-level embeddings. We use a pretrained language model bidirectional encoder representations from transformers (BERT) as the lower-level transformer, which is equivalent to introducing external data into the model and solve the problem of data shortage to some extent. In addition, we add speaker embeddings to the model for the first time, which enables our model to capture the interaction between speakers. Experiments on three dialog emotion datasets, Friends, EmotionPush, and EmoryNLP, demonstrate that our proposed hierarchical transformer network models achieve 1.98%, 2.83%, and 3.94% improvement, respectively, over the state-of-the-art methods on each dataset in terms of macro-F1.

## 1 Introduction

Sentiment analysis, considered one of the most important methods for analyzing real-world communication, is a kind of classification task for extracting emotion from language. It can help us progress in many fields, such as data mining and developing empathetic machines for people. In this paper, we consider one of the tasks in this research direction, utterance-level emotion recognition (ULER) [Poria *et al.*, 2017]. In ULER, an utterance [Olson, 1977] is a unit of speech bounded by breathes or pauses, and its goal is to tag each utterance in a dialog with the indicated emotion (e.g., *happy*, *sad*, or *angry*). Traditional sentiment analysis methods are confined to analyzing only a single sentence or document, regardless of its surrounding information. However, in the field of ULER, contextual information is indispensable in emotional discrimination. For example, in Figure 1, the utterance "*Yes, I agree with this point.*" can deliver different emotions in different contexts. To identify a speaker's emotion precisely, [Hazarika *et al.*, 2018] produced contextual representations for prediction with a recurrent neural network (RNN), where each utterance is represented by a feature vector extracted by convolutional neural networks (CNN) at an earlier stage. Similarly, [Jiao *et al.*, 2019] proposed a hierarchical gated recurrent unit (HiGRU) framework with a lower-level GRU to model the word-level inputs and an upper-level GRU to capture the contexts of utterance-level embeddings. Theoretically, RNNs such as long short-term memory (LSTM) and gated recurrent units (GRUs) should propagate long-term contextual information. However, in practice, this is not always the case [Bradbury *et al.*, 2017]. In cases where the input sequence is long, RNNs may experience an exploding gradient or vanishing gradient. Unlike traditional text classification problems, in the field of ULER, there are a limited number of datasets, and most datasets contain inadequate conversations. This issue limits the possibility of obtaining larger models for this task. To solve this issue, [Zhong *et al.*, 2019] proposed a knowledge-enriched transformer (KET) to effectively incorporate contextual information and external knowledge bases, but this model structure is complex, and the running speed is not high. [Jiao *et at*., 2019] proposed pretraining a context-dependent encoder (CoDE) for ULER by learning from unlabeled conversation data to address the aforementioned challenge, but the model did not perform better in the word-level embedding phase.

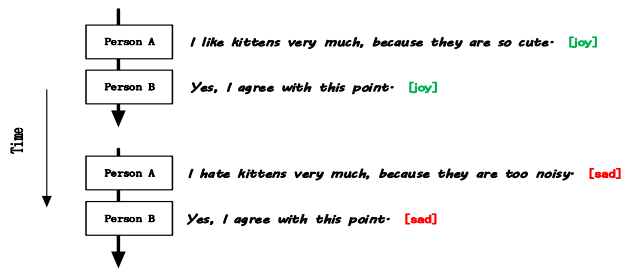

Figure 1: The utterance "*Yes, I agree with this point.*" can deliver different emotions in different contexts.

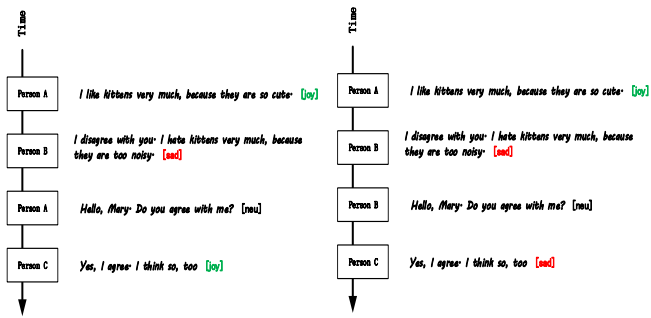

Figure 2: The utterance *"Yes, I agree. I think so, too."* delivers different emotions, *joy* and *sadness*, when the previous sentence is from *Person A* and *Person B*, respectively.

In this task, we propose a hierarchical transformer framework to solve the above issues. First, we use a transformer [Vaswani *et at*., 2017] to model the word-level input and capture the contexts of utterance-level embeddings, which has been shown to be a powerful representation learning model in many NLP tasks and can exploit contextual information more efficiently than RNNs and CNNs. Second, for the data scarcity issue, we use a pretrained language model, bidirectional encoder representations from transformers (BERT) [Devlin *et al*., 2018] as the lower-level transformer, which is equivalent to introducing external data into the model and helps our model obtain better utterance embedding. Third, the same utterance can deliver different emotions in the same context. For example, in Figure 2, the utterance *"Yes, I agree. I think so, too."* can deliver different emotions, *joy* and *sadness*. However, previous studies have not addressed this situation because those models did not capture the interaction between the speakers, and did not consider the emotional dynamics of the speakers in a dialog. To solve the problem, we introduce speaker embedding into our model. To the best of our knowledge, this is the first model for ULER with speaker embedding. After obtaining the contextual utterance embedding vectors with a hierarchical transformer framework, we feed them into the fully connected layers for classification. We employ dropout on the fully connected layers to prevent overfitting. Finally, we obtain an utterance category with a softmax layer.

We summarize our contributions as follows:
• We propose a hierarchical transformer framework to better learn both the individual utterance embeddings and the contextual information of utterances.
• We use a pretrained language model, BERT, to obtain better dialog embedding, which is equivalent to introducing external data into the model and solve the problem of data shortage to some extent.
• For the first time, we use speaker embedding in the model for the ULER task, which allows our model to capture the interaction between speakers and better understand emotional dynamics in dialog systems.
• Our model outperforms state-of-the-art models on three benchmark datasets, Friends, EmotionPush, and EmoryNLP.

## 2 Related work

Text-based emotion recognition is a long-standing research topic, and there have been many excellent studies. However, these models do not perform well in the field of ULER because they treat texts independently and thus cannot capture the interdependence of utterances in dialogs. To capture the contexts of utterance-level embeddings more effectively, we propose a hierarchical transformer framework, which is mainly explored in the following topics.

### 2.1 Individual Utterance Information Extraction

In traditional methods, a common method of expressing text is the bag-of-words method. However, the bag-of-words method loses the order of the words. The n-gram model is a very popular statistical language model and usually performs well [Thorsten, 1998]. However, the n-gram model has a large defect in that it is affected by data sparsity [Bengio *et at*., 2013]. Recently, neural network methods have become increasingly popular. There is a trend moving from traditional methods to deep learning methods to obtain better text representations. Some prominent models include recursive auto-encoders (RAEs) [Socher *et al*., 2011], convolutional neural networks (CNNs) [Kim, 2014], and recurrent neural networks (RNNs) [Abdul-Mageed and Ungar, 2017]. Although we can train a more complex model with a neural network, when the quantity of data is small, it does not perform well.

### 2.2 Pretrained Language Models

Unsupervised pretraining is a special case of semisupervised learning where the goal is to find a good initialization point. Pretrained language models, such as ELMo [Peters *et al*.,2018], OpenAI GPT [Radford *et al*., 2018], and BERT [Devlin *et al*., 2018], have achieved great success in a variety of NLP tasks, such as sentiment analysis and textual classification. They can generate deep contextualized embeddings since they are pretrained on a massive unlabeled corpus (i.e., English Wikipedia). Some proposed models [Sun *et al*., 2019] with pretrained language models have obtained outstanding results on the sentiment analysis task of individual sentences. [Reimers *et at*., 2019] proposed Siamese BERT-networks (SBERT) to obtain sentence embeddings and proved that their model outperforms other state-of-the-art sentence embedding methods.

### 2.3 Contextual Information Extraction

The RNN architecture is a standard method for capturing the sequential relationship of data. [Poria *et at*., 2015] captured the contextual information with a bidirectional long short-term memory (BiLSTM) network and obtained great performance. Similarly, [Jiao et al., 2019] applied bidirectional GRU to model contextual information. In addition, they placed a self-attention layer in the hidden states of GRU and fused the attention outputs with the individual utterance embeddings to learn the contextual utterance embeddings. [Luo et al., 2019] applied self-attention to model the context of textual features extracted by BiLSTM. [Zahiri and Choi, 2018] proposed sequence-based convolutional neural networks (SCNN) that utilize emotion sequences from previous utterances to detect the emotion of the current utterance.

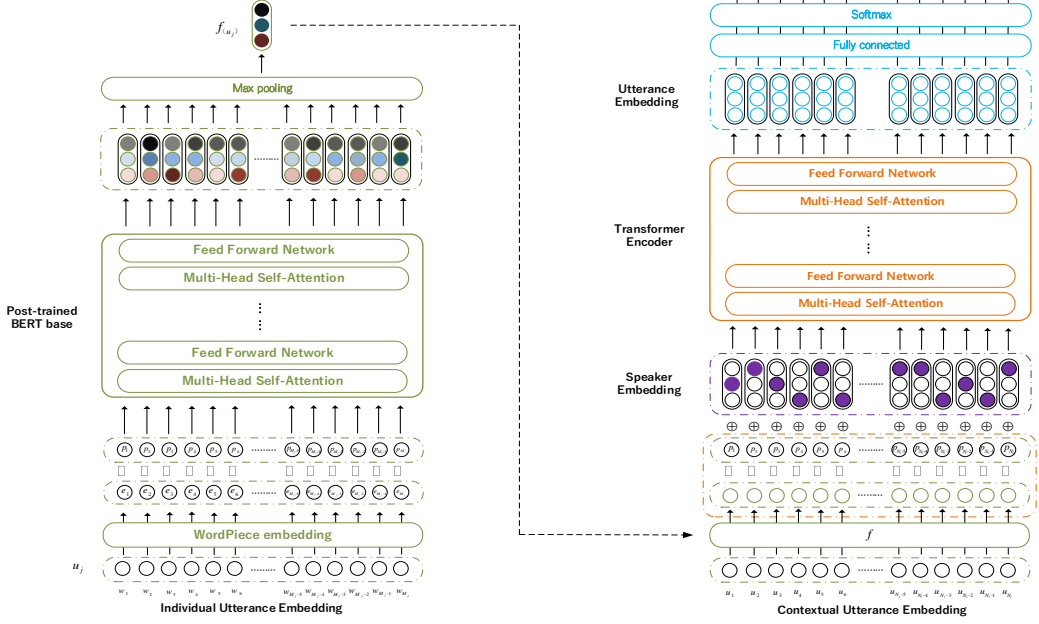

Figure 3: The architecture of our proposed HiTransformer-s. By removing the "Speaker Embedding" layer, we attain HiTransformer.

## 2.4 Transformer

The transformer learns the dependencies between words based entirely on self-attention without any recurrent or convolutional layers. Due to its rich representation and fast computation, it has been applied to many NLP tasks, e.g., response matching in dialog systems [Zhou et al., 2018] and language modeling [Dai et al.,2019]. The success of transformer has raised a large body of follow-up work. Therefore, some transformer variations have also been proposed, such as GPT [Radford et al., 2018], BERT [Devlin et al., 2018], universal transformer [Dehghani et al., 2018] and CN3 [Liu et al., 2018].

## 3 Approach

In this section, we present the task definition and our proposed hierarchical transformer (HiTransformer) network. In addition, we propose a variation in HiTransformer by adding speaker embedding, named HiTransformer-s. The overall architecture of our models is illustrated in Figure 3.

### 3.1 Task Definition

Let there be a set of speakers, $S = \{s_i\}_{i=1}^{M}$, where $M$ is the number of speakers, and a set of emotions, $C = \{c_i\}_{i=1}^{N}$, where $N$ is the number of emotions, such as *anger*, *joy*, *sadness*, and *neutral*. Assume we are given a set of dialogs, $D = \{D_i\}_{i=1}^{L}$, where $L$ is the number of dialogs. In each dialog, $D_i = \{(u_j, s_j, c_j)\}_{j=1}^{N_i}$ is a sequence of utterances, where the utterance $u_j$ is spoken by $s_j \in S$ with an emotion $c_j \in C$. Our goal is to train a model to find the most likely emotion from $C$ for each new utterance.

### 3.2 HiTransformer: Hierarchical Transformer

Our HiTransformer consists of two-level transformers: the lower-level transformer models the word-level input and obtains the individual utterance embedding. The upper-level transformer captures the contextual information and obtains utterance-level embeddings.

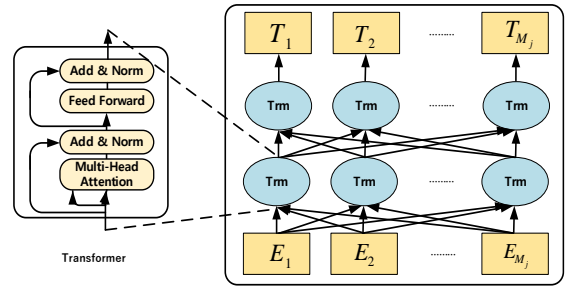

Figure 4: Structure of BERT

**Individual Utterance Embedding**

For the input utterance $u_j = \{w_k\}_{k=1}^{M_j}$, where $u_j$ is the $j-th$ utterance in $D_i$ and $M_j$ is the number of words in the utterance $u_j$. First, the utterance $u_j$ is lower-cased and tokenized according to a byte pair encoding (BPE) algorithm. If there are tokens exceeding the preset maximum length of input tokens, those tokens are excluded from the list. Then, we embed those tokens through WordPiece embeddings [Wu *et at.*, 2016] and obtained the token embeddings $e = \{e_k\}_{k=1}^{M_j}$. Finally, the input embeddings $E = \{E_k\}_{k=1}^{M_j}$ are the summation of the token embeddings $e$ and the positional embeddings $p = \{p_k\}_{k=1}^{M_j}$:

$$E_k = e_k \odot p_k \quad (k \epsilon [1, M_j]) \qquad (1)$$

where $\odot$ denotes $element-wise$ addition.

We feed the input embeddings $E$ into the lower-level transformer to learn the individual utterance embedding. We adopt the transformer-based pretrained language model BERT (illustrated in Figure 4) as the lower-level transformer, which is designed to pretrain deep bidirectional representations from unlabeled text by jointly conditioning both the left and right contexts in all layers. The detailed structure is shown in Figure 4. The language model converts input embeddings $E$ into contextual word embedding $T = \{T_k\}_{k=1}^{M_j}$.

$$T = BERT(E) \qquad (2)$$

The individual utterance embedding is then obtained by max-pooling on the contextual word embeddings within an utterance, which can assist in retaining important information in each dimension:

$$f(u_j) = maxpool(T) \qquad (3)$$

**Contextual Utterance Embedding**

For the $i-th$ dialog in $D$, $D_i = \{(u_j, s_j, c_j)\}_{j=1}^{N_i}$, the individual utterance embedding is $\{f(u_j)\}_{j=1}^{N_i}$. We concatenate the individual embeddings with the position embeddings to obtain $U = \{f(u_j) \odot p_j\}_{j=1}^{N_i}$, where $p_j$ is the embedding of position $j$. Then, we feed $U$ into the upper-level transformer to capture the sequential and contextual relationship of utterances in a dialog and obtain the contextual utterance embedding $t = \{t_j\}_{j=1}^{N_i}$.

$$t = Transformer(U) \qquad (4)$$

Then, we feed the contextual utterance embedding vector into the classifier, which consists of two linear layers, one activation function and dropout. Finally, we obtain the predicted vector over all emotions with a softmax function.

$$selu(x) = \lambda \begin{cases} x & if \ x > 0 \\ \alpha e^x - \alpha & if \ x \leq 0 \end{cases} \qquad (5)$$

$$\hat{y} = softmax(W_2 selu(W_1 t + b_1) + b_2) \qquad (6)$$

### 3.3 HiTransformer-s: Hierarchical Transformer with speaker embeddings

The HiTransformer contains a main issue that it cannot capture the interaction of speakers in a dialog. For example, in Figure 2, the utterance "*Yes, I agree. I think so, too.*" delivers different emotions, *sadness* and *joy*. However, the HiTransformer cannot tag it exactly. To solve this problem, we propose hierarchical transformer with speaker embeddings (HiTransformer-s), which can model the interaction of speakers in a dialog.

For the $i-th$ dialog in $D$, $D_i = \{(u_j, s_j, c_j)\}_{j=1}^{N_i}$, the individual utterance embedding is $\{f(u_j)\}_{j=1}^{N_i}$ and $N_s(D_i)$ is the number of speakers in $D_i$. We first encode all the speakers in $D_i$ with one-hot encoding and then pad them to the dimension of $Max\{N_s(D_i)\}_{i=1}^{L}$ with 0 to obtain the speaker embeddings $\{e_s(s_j)\}_{j=1}^{N_i}$.

$$\{e_s(s_j)\}_{j=1}^{N_i} = pad(onehot(\{s_j\}_{j=1}^{N_i}), Max\{N_s(D_j)\}_{j=1}^{L}) \quad (7)$$

Finally, we concatenate the summation of the individual utterance embeddings and the embeddings of position with the speaker embeddings of every utterance as the input of the upper-level transformer.

$$U = \{(f(u_j) \odot p_j) \oplus e_s(s_j)\}_{j=1}^{N_i} \qquad (8)$$

Where $\odot$ denotes $element-wise$ addition, and $\oplus$ is the concatenation operator.

### 3.4 Model Training

To solve the issue of class imbalance, following the above research [Khosla, 2018], we use weighted cross entropy as the training loss to weight the samples of minority classes as below.

$$loss = -\frac{1}{\sum_{i=0}^{L} N_i} \sum_{i=1}^{L} \sum_{j=1}^{N_i} \frac{1}{w_{c_j}} \sum_{c \in C} y_j^c \log_2(\hat{y}_j^c) \qquad (9)$$

$$w_c = \frac{a_c}{\sum_{i \in C} a_i} \qquad (10)$$

where $a_i$ denotes the number of utterances with emotion $i$ in the training set.

## 4 Experimental Settings

In this section, we present the datasets, evaluation metrics, baselines and experimental results of our model.

### 4.1 Dataset

**Friends** [Hsu and Ku, 2018]: The dataset is annotated from the Friends TV Scripts, and each dialog in the dataset consists of a scene of multiple speakers. In total, there are 1,000 dialogs, which are split into three parts: 720 for training, 80 for validation, and 200 dialogs for testing. Each utterance is tagged with an emotion in a set of emotions, {*anger, joy, sadness, neutral, surprise, disgust, fear,* and *nonneutral*}.

| Dataset | #Dialog (#Utterance) | | | Emotion | | | | |
|---|---|---|---|---|---|---|---|---|
| | Train | Val | Test | Ang | Hap/Joy | Sad | Neu | Others |
| **Friends** | 720(10651) | 80(721) | 200(1208) | 756 | 1710 | 498 | 6530 | 5006 |
| **EmoryPush** | 720(10733) | 80(1202) | 200(2807) | 140 | 2100 | 514 | 9855 | 2133 |

Table 1: Detailed descriptions of Friends and EmotionPush

| Dataset | #Dialog (#Utterance) | | | Emotion | | | | | | |
|---|---|---|---|---|---|---|---|---|---|---|
| | Train | Val | Test | Neutral | Joyful | Peaceful | Powerful | Scared | Mad | Sad |
| **EmoryNLP** | 713 (9934) | 99 (1344) | 85 (1328) | 3776 | 2755 | 1191 | 1063 | 1645 | 1332 | 844 |

Table 2: Detailed descriptions of EmoryNLP

**EmotionPush** [Hsu and Ku, 2018]: The dataset consists of private conversations between friends on Facebook include 1,000 dialogs, which are split into 720, 80, and 200 dialogs for training, validation and testing, respectively. Each utterance is tagged with an emotion in a set of emotions as in the Friends dataset.

**EmoryNLP** [Zahiri and Choi, 2018]: The dataset is annotated from the Friends TV Scripts as well. It includes 713 dialogs for training, 99 dialogs for validation and 85 dialogs for testing. The emotion labels include *neutral*, *sad*, *mad*, *scared*, *powerful*, *peaceful*, and *joyful*.

For the first two datasets, we follow previous works [Jiao et at., 2019] to consider only four emotion classes, i.e., *anger*, *joy*, *sadness*, and *neutral*, and consider all the emotion classes for EmoryNLP.

### 4.2 Evaluation Metrics

Following [Jiao et at., 2019], which achieved the best performance on several ULER datasets, we choose macro-averaged F1-score as the primary metric for evaluating the performance of our models.

$$macro - F1 = \frac{\sum_{c \in C} F1_c}{|C|} \quad (11)$$

where $F1_c$ is the F1-score of emotion $c$. We also report the weighted accuracy (WA) and unweighted accuracy (UWA), which were adopted in a previous work [Hsu and Ku, 2018].

$$WA = \sum_{c \in C} w_c a_c \quad (12)$$
$$UWA = \frac{\sum_{c \in C} a_c}{|C|} \quad (13)$$

where $w_c$ is the percentage of class $c$ in the testing set, and $a_c$ is the corresponding accuracy. As shown in Table 1 and Table 2, most of the datasets in this paper have an imbalanced emotion distribution, so the F1-score is better for measuring the model performance.

### 4.3 Compared Methods

We compare our model HiTransformer and HiTransformer-s with the following state-of-the-art baselines:

**SA-BiLSTM** [Luo et at., 2018]: A self-attentive bidirectional LSTM model, an efficient model that achieved second place in the EmotionX Challenge [Hsu and Ku, 2018];

**CNN-DCNN** [Khosla, 2018]: A convolutional-deconvolutional autoencoder with more handmade features, and the winner of the EmotionX Challenge [Hsu and Ku, 2018];

**bcLSTM$_+$** [Jiao et at., 2019]: A model with a 1-D CNN to extract the utterance embeddings, and a bidirectional LSTM to model the relationship of utterances;

**bcGRU** [Jiao et at., 2019]: A variant of **bcLSTM$_+$** with a BiGRU to capture the utterance-level context;

**CoDE$_{mid}$** [Jiao et at., 2019]: **CoDE$_{mid}$** is a context-dependent encoder (CoDE) model with a bidirectional GRU that extracts the utterance embeddings and a bidirectional GRU that models the relationship of utterances;

**PT − CoDE$_{mid}$** [Jiao et at., 2019]: A variant of **CoDE$_{mid}$** that pretrains a context-dependent encoder (CoDE) for ULER by learning from unlabeled conversation data;

**HiGRU** [Jiao et al., 2019]: A hierarchical gated recurrent unit (HiGRU) framework with a lower-level GRU to model the word-level inputs and an upper-level GRU to capture the contexts of utterance-level embeddings;

**HiGRU-f** [Jiao et al., 2019]: A variant of HiGRU with individual feature fusion;

**HiGRU-sf** [Jiao et al., 2019]: A variant of HiGRU with self-attention and feature fusion;

**SCNN** [Zahiri and Choi, 2018]: A sequence-based convolutional neural networks that utilizes the emotion sequence from the previous utterances for detecting the emotion of the current utterance.

### 4.4 Parameters

We adopt the pretrained *uncased BERT-Base*[1] model as the lower-level transferable language model, where the maximum input length is 512. The number of combination layers of a multi-head attention and a feedforward neural network is 12. For the upper-level transformer layers, the number of transformer layers is 4 and the number of heads in the multi-head attention is 8. For the classification layer, the internal hidden size of the classification layer is set to 300, and the dropout rate is 0.5 to prevent overfitting. We adopt Adam [Kingma and Ba, 2015] as the optimizer with a batch size of 1 and a learning rate of $1 \times 10^{-5}$.

---
[1]https://github.com/huggingface/transformers

| Model | Friends | | | EmotionPush | | | EmoryNLP | | |
|---|---|---|---|---|---|---|---|---|---|
| | Macro-F1 | WA | UWA | Macro-F1 | WA | UWA | Macro-F1 | WA | UWA |
| SA-BiLSTM | - | 79.8 | 59.6 | - | **87.7** | 55.0 | - | - | - |
| CNN-DCNN | - | 67.0 | 62.5 | - | 75.7 | 62.5 | - | - | - |
| **bcLSTM$_+$** | 63.1 | 79.9 | 63.3 | 60.3 | 84.8 | 57.9 | 25.5 | 33.5 | 27.6 |
| bcGRU | 62.4 | 77.6 | 66.1 | 60.5 | 84.6 | 56.9 | 26.1 | 33.1 | 27.4 |
| **CoDE$_{mid}$** | 62.4 | 78.0 | 65.3 | 60.3 | 84.2 | 58.5 | 26.7 | 34.7 | 28.8 |
| **PT − CoDE$_{mid}$** | 65.9 | 81.3 | 66.8 | 62.6 | 84.7 | 60.4 | 29.1 | 36.1 | 30.3 |
| HiGRU | - | 74.4 | 67.2 | - | 73.8 | 66.3 | - | - | - |
| HiGRU-f | - | 71.3 | 68.4 | - | 73.0 | 66.9 | - | - | - |
| HiGRU-sf | - | 74.0 | **68.9** | - | 73.0 | **68.1** | - | - | - |
| SCNN | - | - | - | - | - | - | 26.9 | 37.9 | - |
| HiTransformer | 66.66 | 82.11 | 63.71 | 63.90 | 86.87 | 61.55 | 31.36 | 37.25 | 29.24 |
| HiTransformer-s | **67.88** | **82.18** | 68.78 | **65.43** | 86.92 | 63.03 | **33.04** | **37.98** | **32.67** |

Table 3: Testing results on Friend, EmotionPush, and EmoryNLP

## 5 Result Analysis

We report the empirical results in Table 3, which present the overall performance of our models on all datasets. From these results, we make the following observations.

### 5.1 Comparison with Baselines

Our proposed HiTransformer-s outperforms the state-of-the-art methods with significant margins on all the datasets in terms of macro-F1 score. Specifically, HiTransformer-s obtains 1.98%, 2.83%, and 3.94% absolute improvement on Friends, EmotionPush, and EmoryNLP, respectively. In addition, for Friends, HiTransformer-s obtains 0.88% improvement compared with the best performance in the past in terms of WA, and 0.12% less than the best performance from HiGRU-sf in terms of UWA. However, HiTransformer-s obtains an 8.18% improvement compared with HiGRU-sf in terms of WA. For EmotionPush, although HiTransformer-s is 0.78% lower than SA-BiLSTM in terms of WA, HiTransformer-s is 8.03% above SA-BiLSTM in terms of WA. Similarly, HiTransformer-s is 5.07% lower than HiGRU-sf in terms of UWA and 13.92% above HiGRU-sf in terms of WA. For EmoryNLP, HiTransformer-s obtains 1.88% and 2.37% absolute improvement in terms of WA and UWA, respectively. The HiTransformer outperforms the state-of-the-art methods on all the datasets in terms of the macro-F1 score as well. The above results demonstrate the superior power of HiTransformer-s and HiTransformer.

### 5.2 HiTransformer vs. HiTransformer-s

By analyzing ULER, we find that speaker information plays an important role in utterance classification. Therefore, we proposed HiTransformer-s on the basis of HiTransformer. From Table 3, we observe that HiTransformer-s outperforms HiTransformer on all three datasets in terms of macro-F1, WA, and UWA. Specifically, on Friends, HiTransformer-s attains 1.22%, 0.07% and 5.07% improvement over HiTransformer in terms of macro-F1, WA, and UWA, respectively.

On EmotionPush, HiTransformer-s attains 1.53%, 0.05% and 1.52% improvement over HiTransformer in terms of macro-F1, WA, and UWA. On EmoryNLP, HiTransformer-s attains 1.68%, 0.73% and 3.43% improvement over HiTransformer in terms of macro-F1, WA, and UWA. The results demonstrate that HiTransformer-s including speaker information is indeed capable of boosting the performance of the HiTransformer.

## 6 Conclusion

In this work, to address utterance-level emotion recognition in dialog systems, we propose a hierarchical transformer (HiTransformer) framework with a lower-level transformer to model word-level input and an upper-level transformer to capture the contexts of utterance-level embeddings. To obtain better individual utterance embeddings, we adopt BERT, which is pretrained on a massive unlabeled corpus as the lower-level transformer. To enable HiTransformer to obtain speaker information, we propose HiTransformer-s. Our proposed hierarchical transformer models outperform the state-of-the-art methods on all three datasets, which demonstrates that hierarchical transformer models can sufficiently capture the available utterance information in a dialog. In the future, we plan to pretrain a transformer model to capture the relationship of utterances, similar to BERT, and adopt it as the upper-level transformer to capture the textual information more sufficiently, which can also address the problem of data scarcity in ULER.